\pdfoutput=1
\documentclass[11pt]{article}

\usepackage[preprint]{acl}

\usepackage{times}
\usepackage{latexsym}
\usepackage[T1]{fontenc}
\usepackage[utf8]{inputenc}
\usepackage{microtype}
\usepackage{inconsolata}
\usepackage{graphicx}

\usepackage{amsmath}
\usepackage{newtxmath}   %
\usepackage{bm}
\usepackage{booktabs}
\usepackage{multirow}
\usepackage{makecell}
\usepackage{pifont}
\usepackage{xspace}
\usepackage{enumitem}
\usepackage{colortbl}
\usepackage{tabularx}
\usepackage{array}
\usepackage{placeins}
\usepackage{float}
\usepackage{algorithm}
\usepackage{algpseudocode}

\setlist[itemize]{leftmargin=*,itemsep=1pt,topsep=2pt,parsep=0pt}

\newcommand{\method}{\textup{\textsc{Grasp}}\xspace}
\newcommand{\rlname}{\mbox{GSPO-EGC}\xspace}
\newcommand{\dataname}{\textup{MedTerse}\xspace}
\newcommand{\cmark}{\ding{51}}
\newcommand{\xmark}{\ding{55}}

\newcommand{\dd}[1]{{\scriptsize\,(#1)}}

\newcommand{\ra}{$\rightarrow$}

\definecolor{oursrow}{HTML}{FCEBEF}
\definecolor{oursblue}{HTML}{C23E62}
\definecolor{graytext}{HTML}{6B7280}

\newcommand{\ourscolor}{\rowcolor{oursrow}}
\newcommand{\codeurl}{\href{https://github.com/sea-comet/GRASP-MedVQA}{\texttt{github.com/sea-comet/GRASP-MedVQA}}}
\newcommand{\dataurl}{{\def\UrlBreaks{\do\/}%
  \href{https://huggingface.co/datasets/seacomet/medterse-vqa}{\nolinkurl{huggingface.co/datasets/seacomet/medterse-vqa}}}}

\title{\method: Group-Adaptive Hybrid RL-SFT Post-Training\\for Medical VQA without Reasoning Annotations}

\author{
  \begin{tabular}{@{}c@{\hspace{1.6em}}c@{\hspace{1.6em}}c@{\hspace{1.6em}}c@{}}
    \textbf{Wanqi Dong\textsuperscript{1}} & \textbf{Wanyue Dong\textsuperscript{2}} &
    \textbf{Jingze Ge\textsuperscript{1}} & \textbf{Mehul Motani\textsuperscript{1}} \\
    \small\mdseries wanqi.dong@u.nus.edu & \small\mdseries dongwanyue166@gmail.com &
    \small\mdseries jingze.ge@u.nus.edu & \small\mdseries motani@nus.edu.sg
  \end{tabular}
\\[4pt]
  \normalsize
  \textsuperscript{1}National University of Singapore,
  \textsuperscript{2}Shandong Provincial Daizhuang Hospital
}

\begin{document}
\maketitle

\begin{abstract}
Adapting general-purpose vision-language models (VLMs) to medical visual question answering (VQA) is valuable for helping clinicians read images and make diagnoses, but annotated data are scarce and provide only short answers, without reasoning annotations. Supervised fine-tuning (SFT) fits these labels but overfits to them and amplifies their biases, eroding out-of-domain (OOD) generalization. Reinforcement learning (RL) better preserves generalization but never uses the gold answer as a training target: GRPO-style algorithms learn from the contrast between correct and incorrect answers within a group of sampled responses, so all-correct or all-wrong groups give no correctness signal. Existing SFT-RL hybrids either apply SFT and RL in separate stages or switch between them per question, or need reasoning traces from experts or stronger models, which medical VQA data lack.
We propose \method, which combines RL and SFT in every update using only short answers: RL lets the model explore its own reasoning, SFT teaches the gold answer after that reasoning to supply the direct supervision RL lacks, and their balance adapts to how well the model answers each question.
Tuning only about 1\% of the parameters with LoRA, \method outperforms all compared training objectives both in domain and on OOD open-ended questions across Qwen3-VL-8B, InternVL3-8B and Gemma~3-4B. On an OOD benchmark spanning eight imaging modalities, \method-trained InternVL3-8B and Qwen3-VL-8B beat all 10 compared general and medical VLMs in open-ended accuracy, including models trained on up to 2,600 times more data or with over 250 times more trainable parameters. We release \dataname, a dataset of 12.9K short-answer questions, and our code at \codeurl.
\end{abstract}

\vfill
\section{Introduction}
\label{sec:intro}

\begin{figure*}[t]
    \centering
    \includegraphics[width=\textwidth]{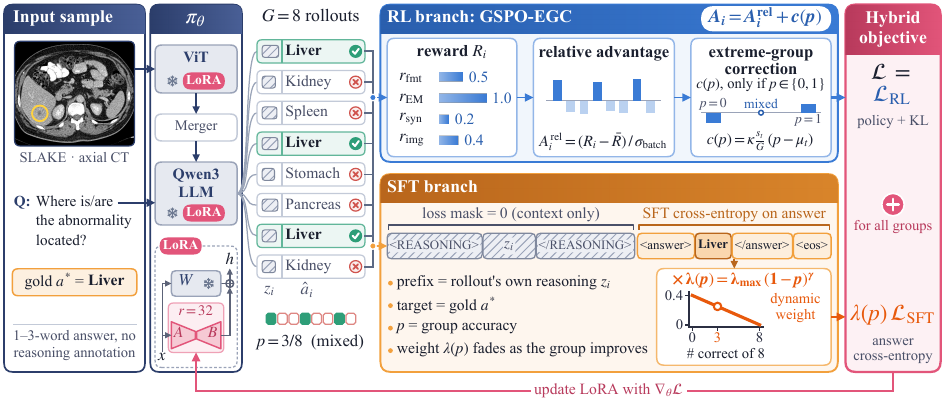}
    \caption{\textbf{\method: simultaneous RL and SFT training on the same rollouts.} For each question with only a short gold answer, the policy samples $G{=}8$ reason-then-answer rollouts whose group accuracy is $p$. \emph{Top (RL):} group-relative advantages plus an adaptive extreme-group correction $c(p)$ for all-wrong and all-correct groups drive the \rlname loss. \emph{Bottom (SFT):} each rollout's own reasoning is kept as a loss-free prefix, followed by the teacher-forced gold answer, weighted by $\lambda(p)$. \emph{Policy:} Qwen3-VL-8B with frozen base weights; both losses update only the LoRA adapters in the ViT and the LLM, and no reasoning annotations are used.}
    \label{fig:method}
\end{figure*}

General-purpose vision-language models (VLMs) show strong image understanding and reasoning. Adapted to medical visual question answering (VQA), they can answer free-form questions about an image, such as which organ is shown or where an abnormality lies, and serve as a second reader that assists clinicians in diagnosis. However, many medical VQA benchmarks~\citep{omnimedvqa}, and the RL-trained medical VLMs built on them~\citep{medvlmr1}, rely on multiple-choice questions, whereas clinical image reading offers no options; without options, current medical VLMs get most questions wrong (\S\ref{sec:res-vlm}). We therefore study open-ended short-answer VQA, a harder and more realistic setting, under two constraints. (1)~\emph{Scarce data:} expert annotation is costly and privacy rules limit data sharing, so annotated datasets are small and their answers are only a few words long, with no reasoning annotations. (2)~\emph{Fit without losing generalization:} a deployed model encounters images from other hospitals, devices and modalities, so post-training must fit the training domain without eroding what the base model knows.

The two standard post-training recipes fall short of these requirements in opposite ways (Table~\ref{tab:main}). \emph{Supervised fine-tuning} (SFT) fits the training domain quickly but tends to overfit and amplify label biases, generalizing poorly out of domain (OOD). \emph{Reinforcement learning with verifiable rewards} (RLVR) algorithms such as GRPO~\citep{shao2024deepseekmath} and GSPO~\citep{zheng2025gspo} better preserve the base model's generalization but learn only from contrasts among the model's own answers; when every sampled answer is wrong, they receive no signal toward the correct answer. Some of the strongest medical VLMs sidestep the data constraint through scale, training all parameters on millions of curated examples (Table~\ref{tab:vlm}), a cost most medical research groups cannot afford.

A few recent works have begun to combine SFT and RL, but most target text-only LLMs and are not fully suited to this setting. Some apply one signal at a time: \emph{staged} training (SFT, then RL) can erode what SFT learned once RL begins, and \emph{switched} training~\citep{lv2025hpt} picks SFT or RL per question, withholding supervision from partly solved questions and RL from unsolved ones. Others imitate expert or distilled reasoning traces~\citep{yan2025luffy,fu2025srft}, which medical VQA datasets rarely provide, and falling back to supervising the bare answer, without reasoning, costs accuracy (\S\ref{sec:ablation}). Adding the bare gold answer to the rollout group (gold injection)~\citep{visurf} teaches the policy to skip its reasoning (\S\ref{sec:why}).

To meet both constraints with short answers alone, we propose \method (\textbf{G}roup-adap\-tive \textbf{R}ein\-force\-ment with \textbf{A}n\-swer \textbf{S}u\-per\-vi\-sion on self-gen\-er\-at\-ed rea\-son\-ing \textbf{P}aths; Figure~\ref{fig:method}). \\
Our contributions are as follows:
\begin{itemize}
    \item \method is a simultaneous RL-SFT hybrid that, on the same rollouts, supervises the gold answer after the model's own reasoning, without reasoning annotations. It outperforms pure SFT, pure RL, and staged, switched and gold-injection hybrids, especially on open-ended questions in domain and OOD, and its advantage holds across backbones.
    \item A group-adaptive weight concentrates supervision on questions the model gets wrong and decays as the model learns to answer them.
    \item The RL branch, \rlname, adaptively corrects the advantages of all-wrong and all-correct groups, which otherwise carry no correctness signal, according to the model's running accuracy.
    \item We release \dataname, 12.9K short-answer questions filtered and rewritten from four public medical VQA datasets, and an open-ended version of OmniMedVQA with eight imaging modalities.
    \item Experiments show that InternVL3-8B and Qwen3-VL-8B trained with \method, tuning only about 1\% of their parameters on 12.9K questions, surpass all 10 general and medical VLMs on OOD open-ended questions, including models trained on up to 2,600 times more data or with over 250 times more trainable parameters (27.3\% vs.\ at most 21.6\% exact match).
\end{itemize}

\section{Related Work}
\label{sec:related}

\paragraph{Medical VQA.}
Adapting general VLMs to medical VQA has followed two main routes. The first is large-scale supervised training, and many models use SFT alone. HuatuoGPT-Vision~\citep{huatuogptvision} rewrites PubMed image-text pairs into VQA data with GPT-4V and follows the LLaVA recipe~\citep{llava15} of alignment followed by instruction tuning. HealthGPT-Pro~\citep{healthgptpro} likewise trains all parameters in these two stages, and Lingshu~\citep{lingshu} first trains only the vision encoder and projector and then all parameters in two further stages, both on millions of medical and general examples. MedGemma~\citep{medgemma} and Fleming-VL~\citep{flemingvl} add RL after such supervised training. Such models are costly to train, and their in-domain gains do not always transfer to new sources (\S\ref{sec:res-vlm}). The second route replaces imitation with RLVR. MedVLM-R1~\citep{medvlmr1}, MedVLThinker~\citep{medvlthinker} and MediX-R1~\citep{medixr1} use GRPO-style RL alone and differ mainly in reward design: the first two reward only the format and the chosen option on multiple-choice questions, whereas MediX-R1 scores open-ended answers by LLM-judged accuracy, medical-embedding similarity, format and modality recognition. \method's reward relies mainly on exact match, with no LLM judge. However, RL-only methods can only reinforce answers the model already samples, and DAPO~\citep{yu2025dapo} over-samples and discards all-wrong and all-correct groups; \method instead keeps them, corrects their advantages and supervises the all-wrong ones with the gold answer, while tuning only parameter-efficient adapters.

\paragraph{Combining SFT and RL.}
To combine the fast in-domain fitting of SFT with the generalization of RL, a few recent works have begun to mix the two signals, differing in how they are combined. The most direct option is staged training, supervision followed by RL, as in MedGemma and Fleming-VL. HPT~\citep{lv2025hpt} casts SFT and RL as instances of one policy-gradient estimator and switches per question, applying SFT on a demonstration trace when rollout accuracy does not exceed a threshold (e.g., all rollouts are wrong) and RL otherwise, so each question receives one signal per step. LUFFY~\citep{yan2025luffy} and SRFT~\citep{fu2025srft}, designed for math reasoning with text-only LLMs, use both signals in one step by adding traces from strong reasoners such as DeepSeek-R1~\citep{guo2025deepseekr1} to the GRPO group as off-policy samples, and SRFT also applies an SFT loss to these traces; both, however, need a full reasoning trace for every question, whereas medical VQA data usually have only short answers. Without reasoning annotations, RFT~\citep{yuan2023rft} trains only on the model's own correct reasoning, so never-solved questions get no supervision. ViSurf~\citep{visurf}, designed for VLMs, scores the gold answer as an extra rollout of the group. Unlike these methods, \method applies both signals to the same rollouts in one update, supervising the gold answer after each rollout's own reasoning, weighted by group accuracy.

\section{Method}
\label{sec:method}

To fit short-answer data and still generalize, \method trains two branches on the same rollouts in every update: the RL branch learns from the model's own reasoning and answers, preserving what the base model knows, and the SFT branch pulls the questions RL fails on toward the gold answer. We walk through one update following Figure~\ref{fig:method}.

\subsection{Group Sampling and Rewards}
\label{sec:sampling}

A training example $(I,q,a^\star)$ is a medical image, a question and a gold answer of one to three words, with no reasoning annotation. The policy $\pi_\theta$ is a VLM with LoRA adapters~\citep{hu2022lora} in its vision encoder and language model (Figure~\ref{fig:method}). Given $x=(I,q)$, it writes a reasoning $z$ in \texttt{<REASONING>} tags and then an answer $\hat a$ in \texttt{<answer>} tags. For each question, $G$ rollouts $y_i=(z_i,\hat a_i)$ of $|y_i|$ tokens are sampled from $\pi_{\theta_{\text{old}}}$, and $\mathrm{EM}_i\in\{0,1\}$ indicates whether $\hat a_i$ matches $a^\star$ after normalization. The reward sums four terms (App.~\ref{app:hparams}):
\begin{equation}
    R_i=1.0\,r_{\text{EM}}+0.2\,r_{\text{syn}}+0.4\,r_{\text{img}}+0.5\,r_{\text{fmt}},
    \label{eq:reward}
\end{equation}
where $r_{\text{EM}}$ scores the normalized exact match between $\hat a$ and $a^\star$ and is the main training signal. $r_{\text{syn}}$ computes the BERTScore~\citep{bertscore} between $\hat a$ and $a^\star$ with Bio\_ClinicalBERT~\citep{clinicalbert_acl} to give partial credit to synonymous answers. $r_{\text{img}}$ measures the similarity between the image and the reasoning with BiomedCLIP~\citep{biomedclip} to encourage reasoning grounded in the image. $r_{\text{fmt}}$ penalizes malformed, repetitive or truncated outputs to preserve the reason-then-answer structure that the SFT branch relies on. Each group is characterized by its accuracy
\begin{equation}
    p=\frac{1}{G}\sum_{i=1}^{G}\mathrm{EM}_i.
    \label{eq:p}
\end{equation}
A group is \emph{all-wrong} if $p{=}0$, \emph{all-correct} if $p{=}1$ and \emph{mixed} otherwise; the first two are called extreme groups. In Figure~\ref{fig:method}, three of the eight rollouts answer \emph{Liver}, so $p{=}3/8$ and the group is mixed. With a binary exact-match reward, extreme groups are common, about four in five in the first epoch.

\subsection{RL Branch: \rlname}
\label{sec:rl}

The RL branch (Figure~\ref{fig:method}, top) learns by comparing answers within a group, a comparison that breaks down in extreme groups; \rlname (EGC: extreme-group correction) addresses this by modifying the advantage. As in GRPO, the group mean reward $\bar R$ serves as the baseline, but the reward is normalized by its batch-level standard deviation, so that extreme groups separated only by auxiliary rewards are not inflated into noise:
\begin{equation}
    A^{\text{rel}}_i=\frac{R_i-\bar R}{\sigma_{\text{batch}}+\varepsilon}.
    \label{eq:arel}
\end{equation}
In an extreme group, all rollouts are equally correct and $A^{\text{rel}}$ retains only auxiliary-reward differences, so the wrong answers of an all-wrong group are not pushed down as a whole and the correct answers of an all-correct group are not reinforced. Dropping such groups, as DAPO does, leaves never-solved questions without any gradient, and a fixed-sign advantage set by correctness~\citep{rlzvp} ignores the model's current skill. \rlname instead compares the group's outcome with the running training accuracy $\mu_t$ and adds the \emph{extreme-group correction} to every rollout's advantage, $A_i=A^{\text{rel}}_i+c(p)$, with
\begin{equation}
    c(p)=\begin{cases}\kappa\,\dfrac{s_t}{G}\,(p-\mu_t), & p\in\{0,1\},\\[4pt] 0, & 0<p<1.\end{cases}
    \label{eq:zvc}
\end{equation}
Here $\mu_t$ and $s_t$ are exponential moving averages of the batch mean group accuracy and of the root mean square of $A^{\text{rel}}$ over training steps $t$. The hyperparameter $\kappa$ sets the correction strength, and the sensitivity analysis in \S\ref{sec:sensitivity} gives $\kappa{=}1$. Intuitively (Table~\ref{tab:division}), when the model usually answers correctly, a group with no correct answer is unusual and is penalized more; when correct answers are still rare, a fully solved group is rewarded more. $s_t$ matches the correction to the typical scale of $A^{\text{rel}}$, and the factor $1/G$ makes the correction summed over a group worth one rollout's advantage; over a batch the two sides largely cancel, leaving a mean near zero (App.~\ref{app:method}). The correction can be viewed as an empirical-Bayes-style correction applied only to extreme groups~\citep{ebpo}.

The RL branch then optimizes GSPO's sequence-level objective with the corrected advantage. Its importance ratio $\rho_i$ is the geometric mean of the token-level ratios over the rollout,
\begin{equation}
\begin{aligned}
    \rho_i&=\bigg(\frac{\pi_\theta(y_i\mid x)}{\pi_{\theta_{\text{old}}}(y_i\mid x)}\bigg)^{1/|y_i|}\\
    &=\exp\bigg(\frac{1}{|y_i|}\sum_{t=1}^{|y_i|}\log\frac{\pi_\theta(y_{i,t}\mid x,y_{i,<t})}{\pi_{\theta_{\text{old}}}(y_{i,t}\mid x,y_{i,<t})}\bigg),
\end{aligned}
    \label{eq:ratio}
\end{equation}
and the RL loss is
\begin{equation}
    \mathcal{L}_{\text{RL}}=-\frac{1}{G}\sum_{i=1}^{G}\min\big(\rho_iA_i,\ \bar\rho_iA_i\big),
    \label{eq:rl}
\end{equation}
where $\bar\rho_i=\mathrm{clip}(\rho_i,1-\epsilon_l,1+\epsilon_h)$. In an all-wrong group, however, the correction only pushes down the sampled wrong answers without revealing the right one; the SFT branch supplies it.

\subsection{SFT Branch: Answer Supervision on Own Reasoning}
\label{sec:selfsft}

The SFT branch brings the gold answer into training. Medical VQA offers no reasoning to imitate, and supervising the bare answer would teach the model to skip reasoning, against the reason-then-answer behavior the RL branch optimizes. We therefore let each rollout's own reasoning serve as the context for supervision (Figure~\ref{fig:method}, bottom), keeping the reasoning $z_i$ and replacing the answer with $a^\star$ (\texttt{<R>} abbreviates \texttt{<REASONING>}):
\begin{equation*}
\small
    \tilde y_i=\big[\underbrace{\texttt{<R>}z_i\texttt{</R>}}_{\text{mask }0}\,\underbrace{\texttt{<answer>}a^\star\texttt{</answer><eos>}}_{\mathcal{M}_i:\ \text{mask }1}\big].
\end{equation*}
The reasoning carries no loss; token-averaged cross-entropy is computed only on the answer span $\mathcal{M}_i$, whether or not the rollout was correct:
\begin{equation}
    \mathcal{L}_{\text{SFT}}=-\frac{1}{M}\sum_{i=1}^{G}\log\pi_\theta(a^\star\mid x,z_i),
    \label{eq:sft}
\end{equation}
where $\log\pi_\theta(a^\star\mid x,z_i)$ sums the log-probabilities of the tokens in $\mathcal{M}_i$ and $M=\sum_{i}|\mathcal{M}_i|$.
This requires no expert or distilled traces, and the answer is learned in a context the model itself produces. Since the reasoning receives no imitation gradient, the model is not trained to repeat its own, possibly wrong, explanation; shaping the reasoning is left to RL. The objective also raises a lower bound on $\log\pi_\theta(a^\star\mid x)$, and the rare unusable reasoning spans are truncated or borrowed (App.~\ref{app:method}).

\subsection{Group-Adaptive Weight and Joint Objective}
\label{sec:weight}

Groups differ in how much supervision they need. In an all-wrong group, the correction only pushes down the sampled wrong answers, so the gold answer has to come from the SFT branch. A mixed group already learns from RL's within-group contrast, and an all-correct group already answers correctly and is reinforced by the correction. The supervision weight therefore decreases with group accuracy:
\begin{equation}
    \lambda(p)=\lambda_{\max}\,(1-p)^{\gamma},
    \label{eq:lambda}
\end{equation}
where the hyperparameters $\lambda_{\max}$ and $\gamma$ control the overall strength and the decay rate. Based on the sensitivity analysis in \S\ref{sec:sensitivity}, we set $\lambda_{\max}{=}0.4$ and $\gamma{=}1$, so a group with $k$ of 8 correct rollouts receives $0.4(1-k/8)$ (Figure~\ref{fig:method}, inset). All-wrong groups thus receive the largest weight and all-correct groups none, so that cross-entropy does not keep sharpening the answer distribution on questions the model already solves. As the right side of Figure~\ref{fig:method} shows, \method minimizes for each group
\begin{equation}
    \mathcal{L}=\frac{\mathcal{L}_{\text{RL}}}{s_{\text{RL}}}+\lambda(p)\,\frac{\mathcal{L}_{\text{SFT}}}{s_{\text{SFT}}}+\beta\,\mathbb{D}_{\text{KL}}\big[\pi_\theta\,\|\,\pi_{\text{ref}}\big].
    \label{eq:total}
\end{equation}
The RL term acts on every group and the supervision term acts concurrently whenever $p<1$, making \method a \emph{simultaneous} hybrid; Table~\ref{tab:division} summarizes the division of labor, and App.~\ref{app:method} gives the batch aggregation. $s_{\text{RL}}$, a moving average of the advantage root mean square, scales the RL term (distinct from $s_t$ in Eq.~\ref{eq:zvc}); $s_{\text{SFT}}$ is the mean answer loss over the first training steps and is then frozen, so $\lambda(p)$ alone sets the initial balance of the two branches, and the supervision term shrinks naturally as answers are learned. The KL term keeps the policy close to the base model $\pi_{\text{ref}}$ with a fixed weight $\beta$, outside both normalizations.

\begin{table}[t]
\centering
\footnotesize
\setlength{\tabcolsep}{3pt}
\begin{tabular}{@{}l l c c@{}}
\toprule
\textbf{Group} & \textbf{RL signal} & $c(p)$ & $\lambda(p)$ \\
\midrule
all-wrong & none & $-\kappa\mu_t s_t/G$ & $\lambda_{\max}$ \\
mixed & right vs.\ wrong & $0$ & $\lambda_{\max}(1{-}p)^{\gamma}$ \\
all-correct & none & $+\kappa(1{-}\mu_t)s_t/G$ & $0$ \\
\bottomrule
\end{tabular}
\caption{\textbf{Division of labor between the two branches.} RL signal: whether $A^{\text{rel}}$ separates right from wrong.}
\label{tab:division}
\end{table}

\section{Experimental Setup}
\label{sec:setup}

\begin{table*}[t]
\centering
\footnotesize
\setlength{\tabcolsep}{2.0pt}
\begin{tabular}{@{}l l c ccccccc cccccc@{}}
\toprule
& & & \multicolumn{7}{c}{\textbf{In domain} (local test splits, 6,600 questions)} & \multicolumn{6}{c}{\textbf{OOD} (OmniMedVQA)} \\
\cmidrule(lr){4-10}\cmidrule(l){11-16}
& & & & & & & & & & \multicolumn{3}{c}{open-ended} & & \\
\cmidrule(lr){11-13}
\textbf{Objective} & \textbf{Type} & \textbf{Ep.} & SLAKE & VQA-RAD & PathVQA & Val & Open & Y/N & All & CT & X-ray & MRI & Open & Y/N\textsubscript{bal} & All \\
\midrule
Zero-shot          & --           & --  & 43.4 & 52.5 & 59.6 & 36.6 & 15.2 & 68.5 & 51.9 & 1.0 & 24.2 & 30.1 & 15.9 & 63.6 & 25.6 \\
\midrule
SFT                & single       & 2   & 71.3 & 54.2 & 66.8 & 47.0 & 32.9 & 76.0 & 62.5 & 19.8 & 34.5 & 31.6 & 22.9 & 47.1 & 21.2 \\
GSPO               & single       & 2   & 59.8 & 55.7 & 66.8 & 44.7 & 26.3 & 75.8 & 60.4 & 18.7 & 37.8 & 31.4 & 22.8 & 60.8 & 28.0 \\
\rlname            & single       & 2   & 64.3 & 56.4 & 67.8 & 47.3 & 29.5 & 77.1 & 62.2 & 18.6 & 42.9 & 31.6 & 24.2 & 61.7 & 29.3 \\
\midrule
SFT\ra\rlname      & staged       & 1+1 & 70.6 & 53.5 & 69.0 & 48.2 & 33.1 & 77.8 & 63.9 & 18.5 & 43.9 & 31.6 & 23.0 & 62.6 & 27.0 \\
HPT switch         & switch       & 2   & 75.2 & 55.0 & 74.6 & 52.2 & 36.2 & 83.5 & 68.7 & 21.8 & 41.4 & 32.0 & 24.2 & \textbf{68.3} & \textbf{29.8} \\
ViSurf             & inject       & 2   & 76.1 & 54.5 & 75.5 & 51.7 & 37.0 & 83.9 & 69.3 & 19.1 & 33.8 & 31.2 & 22.7 & 59.8 & 25.4 \\
RFT                & self         & 2   & 45.4 & 51.8 & 61.5 & 38.6 & 16.7 & 70.4 & 53.7 & 1.1 & 31.1 & 30.5 & 17.0 & 64.5 & 26.4 \\
\ourscolor
\method            & simult.      & 2   & \textbf{78.2} & \textbf{57.4} & \textbf{77.7} & \textbf{55.9} & \textbf{40.0} & \textbf{86.4} & \textbf{71.9} & \textbf{22.0} & \textbf{44.8} & \textbf{33.4} & \textbf{25.5} & 60.4 & 28.6 \\
\bottomrule
\end{tabular}
\caption{\textbf{Training objectives on Qwen3-VL-8B} (EM, \%). All rows share the backbone, data, LoRA, rewards and evaluation protocol. Open/Y/N/All: pooled results; CT-MRI: OOD open-ended subsets; Y/N\textsubscript{bal}: balanced OOD yes/no accuracy. Best in bold; seed variation in App.~Table~\ref{tab:app-seeds}.}
\label{tab:main}
\end{table*}

\begin{figure}[t]
    \centering
    \includegraphics[width=0.8\columnwidth]{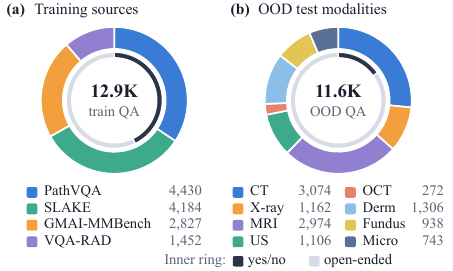}
    \caption{\textbf{Data.} (a) \dataname: 12,893 short-answer training questions from four public sources. (b) 11,575 OOD test questions over eight OmniMedVQA modalities. Inner rings: share of yes/no questions.}
    \label{fig:data}
\end{figure}

\paragraph{Data.}
We build \dataname from the training splits of PathVQA~\citep{pathvqa}, SLAKE~\citep{slake} and VQA-RAD~\citep{vqarad} and the validation items of GMAI-MMBench~\citep{gmaimmbench}, keeping questions with one- to three-word answers. DeepSeek-V4~\citep{deepseekv4} rewrites questions that need their options into direct questions. Rewrites that remain ambiguous are dropped, as are overly broad questions such as \emph{What is present?} The result is 12,893 questions on 5,446 images, none with reasoning annotations (Figure~\ref{fig:data}a). We evaluate in three settings. The \emph{in-domain} set comprises a held-out validation split and the official test splits of the three source datasets (6,600 questions). The \emph{OOD} set converts OmniMedVQA~\citep{omnimedvqa}, a source held out from training, into 11,575 short-answer questions over eight imaging modalities by the same rule, with gold answers copied from option labels corrected (Figure~\ref{fig:data}b). Two \emph{external} sets test endoscopy (Kvasir-VQA;~\citealp{kvasirvqa}) and specific diagnoses (VQA-Med 2020;~\citealp{vqamed2020}). Split statistics are in App.~\ref{app:data}. \dataname and the converted OOD set are publicly available at \dataurl.

\paragraph{Models and training.}
Our main backbone is Qwen3-VL-8B-Instruct~\citep{qwen3vl}, and we also train InternVL3-8B~\citep{internvl3} and Gemma~3-4B~\citep{gemma3} to test generality. We train only rank-32 LoRA adapters in the vision encoder and language model (102.7M parameters, 1.17\% of Qwen3-VL-8B). Each step samples $G{=}8$ rollouts at temperature 1.15 for 12 questions, with learning rate $2{\times}10^{-5}$ and a cosine schedule. All objectives train for two epochs; one \method run on Qwen3-VL-8B takes 81 GPU-hours (about 20 hours on four A100s). The main configuration, SFT and GSPO use three seeds (App.~Table~\ref{tab:app-seeds}). Remaining hyperparameters are in App.~\ref{app:hparams}.

\paragraph{Baselines.}
We compare with two kinds of baselines. The first are seven other training objectives that share the backbone, data, LoRA and rewards and differ only in how supervision and RL are used: three use a single signal (SFT, GSPO and \rlname), and four combine the two signals in other ways (staged, switched, gold-injection and self-training). The second are 10 released general and medical VLMs, evaluated with the same prompt, images, decoding and scorer (App.~\ref{app:baselines}).

\paragraph{Metrics.}
We report normalized exact match (EM) on open-ended, yes/no and all questions (``All''), with balanced accuracy for OOD yes/no questions, 93\% of which are answered \emph{no}. Because EM misses synonyms, we also report accuracy under an LLM judge (DeepSeek-V4.1-Flash) that decides whether a prediction and the gold answer mean the same. Two physicians independently labeled 200 prediction-gold pairs, seeing the image and question but not the judge's verdict; on the pairs where they agreed, the judge matched them 93.9\% of the time (Cohen's $\kappa{=}0.87$), so the LLM judge is a reliable complement to EM (App.~\ref{app:judge}).

\section{Results}
\label{sec:results}

\begin{table*}[t]
\centering
\footnotesize
\setlength{\tabcolsep}{2.9pt}
\begin{tabular}{@{}l rr cc cc cc cc cc cc@{}}
\toprule
& & & \multicolumn{4}{c}{\textbf{In domain}} & \multicolumn{4}{c}{\textbf{OOD} (OmniMedVQA)} & \multicolumn{2}{c}{\textbf{Kvasir}} & \multicolumn{2}{c@{}}{\textbf{VQA-Med}} \\
\cmidrule(lr){4-7}\cmidrule(lr){8-11}\cmidrule(lr){12-13}\cmidrule(l){14-15}
& & & \multicolumn{2}{c}{Open} & \multicolumn{2}{c}{All} & \multicolumn{2}{c}{Open} & \multicolumn{2}{c}{All} & \multicolumn{2}{c}{All} & \multicolumn{2}{c}{All} \\
\cmidrule(lr){4-5}\cmidrule(lr){6-7}\cmidrule(lr){8-9}\cmidrule(lr){10-11}\cmidrule(lr){12-13}\cmidrule(l){14-15}
\textbf{Model} & \textbf{Tuned} & \textbf{Data} & EM & LLM & EM & LLM & EM & LLM & EM & LLM & EM & LLM & EM & LLM \\
\midrule
Qwen3-VL-8B       & --    & --     & 15.2 & 25.4 & 51.9 & 55.1 & 15.9 & 26.9 & 25.6 & 30.9 & 35.6 & 42.6 & 4.8 & 5.1 \\
InternVL3-8B      & --    & --     & 16.6 & 27.4 & 49.7 & 53.3 & 19.2 & 30.1 & 22.2 & 31.5 & 38.7 & 40.7 & 5.1 & 5.7 \\
MedGemma-4B       & 4.3B  & 33.6M  & 20.9 & 25.1 & 53.5 & 54.8 & 17.2 & 24.1 & 25.7 & 31.6 & 20.7 & 21.3 & 6.0 & 6.2 \\
MedGemma-27B      & ${\approx}$27B & --$^\ddagger$ & 26.0 & 33.2 & 55.3 & 57.5 & 15.3 & 21.2 & 20.9 & 25.9 & 33.4 & 34.3 & 8.2 & 8.8 \\
HuatuoGPT-V-7B    & 8.3B  & 1.3M   & 7.1 & 11.3 & 45.3 & 46.7 & 3.0 & 5.1 & 7.0 & 9.5 & 28.6 & 28.7 & 4.8 & 5.1 \\
Lingshu-7B        & 8.3B  & 12.1M  & 29.7 & 33.9 & 53.6 & 55.0 & 18.5 & 27.6 & 26.0 & 33.7 & 39.1 & 40.9 & \textbf{10.8} & \textbf{12.2} \\
HealthGPT-Pro-8B  & 8.8B$^\dagger$ & 13M    & 15.5 & 26.4 & 53.4 & 57.1 & 14.7 & 26.1 & 24.5 & 34.2 & 33.1 & 34.5 & 5.7 & 6.0 \\
Fleming-VL-8B     & 7.9B  & 8.0M   & 36.1 & 39.4 & 71.7 & 72.4 & 16.9 & 19.4 & 28.1 & 30.3 & 32.3 & 32.6 & 5.4 & 6.2 \\
MediX-R1-8B       & 8.8B  & 51K    & 11.7 & 24.9 & 45.4 & 52.4 & 3.3 & 23.2 & 9.9 & 26.7 & 21.9 & 29.3 & 4.5 & 4.5 \\
MedVLThinker-7B   & 8.3B  & 16.5K  & 17.0 & 25.5 & 49.0 & 51.6 & 14.7 & 22.5 & 19.6 & 26.2 & 36.0 & 36.8 & 5.9 & 6.5 \\
\ourscolor
\textbf{\method-Qwen} (Ours) & 0.1B  & 12.9K  & 40.0 & \textbf{42.7} & \textbf{71.9} & \textbf{72.6} & 25.5 & 31.9 & 28.6 & 33.9 & \textbf{45.1} & \textbf{45.9} & 7.1 & 7.4 \\
\ourscolor
\textbf{\method-InternVL} (Ours) & 0.1B & 12.9K & \textbf{40.1} & 42.5 & 71.3 & 72.0 & \textbf{27.3} & \textbf{34.1} & \textbf{31.3} & \textbf{37.0} & 43.8 & 43.9 & 8.2 & 8.8 \\
\bottomrule
\end{tabular}
\caption{\textbf{Comparison with general and medical VLMs}, using one shared prompt (own-prompt results: App.~Table~\ref{tab:app-vlm-own}). EM: exact match; LLM: accuracy when an LLM judges semantic equivalence to the gold answer (App.~\ref{app:judge}); Tuned: trainable parameters; Data: post-training examples. \method-Qwen/-InternVL: Qwen3-VL-8B/InternVL3-8B, 2 epochs. Best in bold. $^\dagger$Inferred from released weights; $^\ddagger$not reported separately.}
\label{tab:vlm}
\end{table*}

\subsection{Comparison with Single-Signal Training}
\label{sec:res-objectives}

We first compare \method on Qwen3-VL-8B with three single-signal objectives (Table~\ref{tab:main}): SFT, which fine-tunes on the bare gold answer; GSPO, standard group-relative RL; and \rlname, the RL branch alone. None of them succeeds on both fronts. SFT fits the training domain but loses much of the base model's balance on OOD yes/no questions, with balanced accuracy dropping from 63.6\% to 47.1\%. The two RL objectives largely keep this balance but gain far less in domain, since a group with no correct answer reveals nothing about the gold answer. \method, in contrast, is best on every in-domain split and on OOD open-ended questions, overall and per modality. Table~\ref{tab:main} lists three of the larger OOD modalities; the pooled OOD columns cover all eight. The gains hold across seeds: averaged over three seeds, \method leads SFT and GSPO by 6.4--14.4 points on in-domain and 2.4--2.6 points on OOD open-ended questions, with standard deviations of at most 0.3 (App.~Table~\ref{tab:app-seeds}).

\subsection{Comparison with Hybrid Training}
\label{sec:res-hybrid}

Four further methods combine supervision and reward signals in other ways: the staged SFT\ra\rlname, one epoch of each; the HPT switch, which gives all-wrong groups only answer supervision and all other groups only RL; ViSurf, which injects the gold answer into each group as an extra rollout; and RFT, which self-trains on the base model's own samples that pass the exact-match reward. As Table~\ref{tab:main} shows, each retains only part of the in-domain fit of SFT and the OOD generalization of RL. The staged run regains SFT's in-domain fit but not the OOD open-ended accuracy of \rlname. ViSurf comes close in domain but, like SFT, gains little on OOD open-ended questions, and RFT stays close to the zero-shot model (\S\ref{sec:why}). The HPT switch is the most balanced of the four, yet still trails \method by 3.8 in-domain and 1.3 OOD open-ended points; its only edge, on OOD yes/no questions, gives it a slightly higher OOD overall accuracy.

\subsection{Comparison with Medical VLMs}
\label{sec:res-vlm}

Table~\ref{tab:vlm} compares \method with 10 released VLMs. On OOD open-ended questions, \method on either backbone beats every other model under both EM and the physician-audited LLM judge (up to 27.3\% vs.\ at most 21.6\% EM), and the lead holds on CT, X-ray, MRI and ultrasound separately (App.~Figure~\ref{fig:oodmod}). On these questions \method-Qwen gains 9.6 points over its base model, with up to 2,600 times less post-training data and 40--260 times fewer trainable parameters than the medical VLMs. In domain, \method is on par with the strongest models: it ranks first with the shared prompt, and with each model's own prompt (App.~Table~\ref{tab:app-vlm-own}) only HealthGPT-Pro edges ahead (73.3\% vs.\ 71.9\%), using about 1,000 times the data and over 80 times the trainable parameters, while falling behind out of domain (21.6\% vs.\ 25.5\% OOD open-ended EM).

Larger models gain little in domain: HuatuoGPT-Vision-34B and MedGemma-27B barely differ from their 7B and 4B versions and still trail \method there by 16--18 points (App.~Table~\ref{tab:app-vlm-own}). MediX-R1, which trains the same backbone with RL alone, trails by over 20 points both in domain and on OOD open-ended questions. The external sets agree: \method beats every model on the held-out Kvasir-VQA endoscopy set under both prompts, and on VQA-Med 2020, where almost every question asks for a difficult specific diagnosis, it is second only to Lingshu among models of up to 9B parameters, and Lingshu trains on the similar VQA-Med 2019. In short, a small adapter trained on 12.9K short answers matches the best medical VLMs in domain and surpasses all of them on OOD open-ended questions.

\subsection{Generalization across Backbones}
\label{sec:res-arch}

\begin{table*}[t]
\centering
\footnotesize
\setlength{\tabcolsep}{2.6pt}
\begin{tabular}{@{}l cccccc cccc@{}}
\toprule
& \multicolumn{6}{c}{\textbf{Components}} & \multicolumn{2}{c}{\textbf{In domain}} & \multicolumn{2}{c}{\textbf{Out of domain}} \\
\cmidrule(lr){2-7}\cmidrule(lr){8-9}\cmidrule(l){10-11}
\textbf{Variant} & Simult. & $\lambda(p)$ & Own $z$ & EGC & $r_{\text{syn}}$ & $r_{\text{img}}$ & Open & All & Open & All \\
\midrule
\ourscolor
\method (2 ep.)                       & \cmark & cont.  & \cmark & adapt. & \cmark & \cmark & 40.0 & 71.9 & 25.5 & 28.6 \\
\quad answer only (no own $z$)        & \cmark & cont.  & \xmark & adapt. & \cmark & \xmark & 35.9\dd{$-$4.1} & 69.7\dd{$-$2.2} & 23.1\dd{$-$2.4} & 24.0\dd{$-$4.6} \\
\quad exclusive switch (HPT)          & \xmark & switch & \cmark & fixed  & \cmark & \cmark & 36.2\dd{$-$3.8} & 68.7\dd{$-$3.2} & 24.2\dd{$-$1.3} & 29.8\dd{+1.2} \\
\quad constant $\lambda{=}0.2$        & \cmark & const. & \cmark & fixed  & \cmark & \cmark & 36.6\dd{$-$3.4} & 68.6\dd{$-$3.3} & 24.6\dd{$-$0.9} & 27.0\dd{$-$1.6} \\
\quad w/o EGC                          & \cmark & cont.  & \cmark & \xmark & \cmark & \cmark & 37.3\dd{$-$2.7} & 69.2\dd{$-$2.7} & 24.5\dd{$-$1.0} & 27.6\dd{$-$1.0} \\
\quad EGC on all-wrong only           & \cmark & cont.  & \cmark & $p{=}0$ & \cmark & \cmark & 38.1\dd{$-$1.9} & 69.9\dd{$-$2.0} & 25.2\dd{$-$0.3} & 27.2\dd{$-$1.4} \\
\quad w/o $r_{\text{syn}}$             & \cmark & cont.  & \cmark & adapt. & \xmark & \cmark & 37.5\dd{$-$2.5} & 68.8\dd{$-$3.1} & 25.0\dd{$-$0.5} & 24.2\dd{$-$4.4} \\
\quad w/o $r_{\text{img}}$            & \cmark & cont.  & \cmark & adapt. & \cmark & \xmark & 37.2\dd{$-$2.8} & 69.4\dd{$-$2.5} & 25.0\dd{$-$0.5} & 31.0\dd{+2.4} \\
\bottomrule
\end{tabular}
\caption{\textbf{Ablations} (EM, \%; Qwen3-VL-8B, two epochs); parentheses give the difference to \method. Simult.: RL on every group and concurrent SFT wherever $p<1$; Own $z$: answer after the rollout's own reasoning; $\lambda(p)$: continuous, constant or switch; EGC: adaptive (both extremes or $p{=}0$ only), fixed (constant shift) or none. Seed s.d.\ of \method: 0.3 (in-domain) and 0.2 (OOD) open-ended points.}
\label{tab:ablation}
\end{table*}

\begin{figure}[t]
    \centering
    \includegraphics[width=\columnwidth]{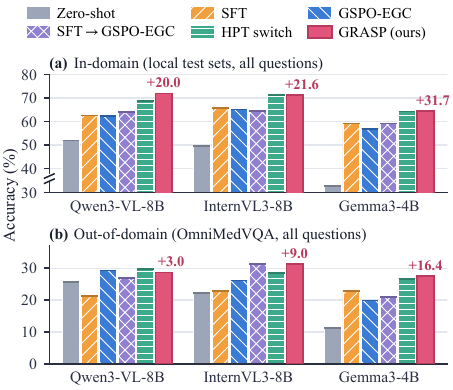}
    \caption{\textbf{Training objectives on three backbones} (EM, all questions; labels: gain of the simultaneous hybrid over zero-shot). OOD open-ended EM on InternVL3-8B: SFT 22.5, \rlname 25.0, SFT\ra\rlname 24.9, \method 27.3.}
    \label{fig:arch}
\end{figure}

Repeating the comparison on InternVL3-8B and the smaller Gemma~3-4B (Figure~\ref{fig:arch}) shows a similar pattern. On InternVL3, \method beats every single-signal and staged control by 5.7--6.9 points in domain and 2.3--4.8 points on OOD open-ended questions. On Gemma~3, \method leads the same controls by 5.2--7.5 points in domain and 4.8--7.8 points out of domain. The hybrid HPT switch also performs well, but remains slightly weaker than \method overall on both backbones: it trails a little in domain and more out of domain (2.8 points on InternVL3 and 1.0 on Gemma~3).

\subsection{Ablations}
\label{sec:ablation}

Table~\ref{tab:ablation} removes one component at a time, and every change costs 1.9--4.1 in-domain open-ended points. The largest drop, 4.1 points in domain and 2.4 on OOD open-ended questions, comes from supervising the bare answer without the rollout's own reasoning (which also removes $r_{\text{img}}$), so the model's own reasoning is a useful context for supervision. How the signals are combined matters as well. The exclusive switch uses almost the same supervision budget as \method yet loses 3.8 points, since mixed groups then receive only RL; a constant SFT weight loses 3.4 points by moving supervision from failed to solved questions. In these runs, where supervision goes matters more than how much. Removing the extreme-group correction costs 2.7 points, and correcting only all-wrong groups recovers part of this loss, so the shift on all-correct groups also helps. Removing $r_{\text{syn}}$ or $r_{\text{img}}$ costs 2.5 and 2.8 points, with OOD open-ended accuracy within 0.5 points. Two variants, the exclusive switch and w/o $r_{\text{img}}$, are slightly higher on OOD overall accuracy (+1.2 and +2.4 points), but only through OOD yes/no questions, and both remain below \method on OOD open-ended questions.

\section{Analysis}
\label{sec:analysis}

\paragraph{Supervision dynamics.}
\label{sec:dynamics}
App.~Figure~\ref{fig:dynamics}a tracks the group types and the supervision weight over the main run. As the model improves, all-correct groups become more common, reaching about half of all groups in the second epoch, while all-wrong and mixed groups become fewer. The supervision weight $\lambda(p)$ averaged over all groups therefore falls from 0.21 to 0.17, and most of it remains on all-wrong groups, where exact match gives no correctness signal. The correction also shifts with training (App.~Figure~\ref{fig:dynamics}b): early on it leans toward rewarding all-correct groups, and as the running accuracy rises it leans toward penalizing all-wrong groups.

\paragraph{Why simultaneous beats the other hybrids.}
\label{sec:why}
In \method, a failing group is pushed away from its wrong answers and toward the gold answer in the same update. In our runs, injecting the bare gold answer into the group, as ViSurf does, teaches the policy to skip reasoning: in all-wrong groups this short sequence receives the largest advantage, and valid reasoning falls from 74\% to under 5\% within 70 steps (App.~\ref{app:baselines}). RFT learns only from questions the model already solves, yet 16 samples never solve nearly three quarters of the open-ended training questions, the questions on which \method concentrates its supervision.

\paragraph{Hyperparameter sweep.}
\label{sec:sensitivity}
App.~Figure~\ref{fig:sensitivity} and Table~\ref{tab:app-sweep} vary $\lambda_{\max}$, $\gamma$ and $\kappa$ one at a time. The default gives the highest in-domain accuracy, while supervision that is too weak ($\lambda_{\max}{=}0.2$) or fades too fast ($\gamma{=}2$) costs 3.8 and 2.6 in-domain open-ended points. OOD open-ended accuracy, by contrast, stays within 0.7 points of the default, so OOD accuracy is insensitive to these hyperparameters in the tested range.

\section{Conclusion}
\label{sec:conclusion}

We presented \method, which applies RL and answer supervision simultaneously to the same rollouts: the gold answer is supervised after the model's own reasoning and weighted by group accuracy, and \rlname corrects the advantages of all-wrong and all-correct groups. \method therefore needs only short answers and no reasoning annotations. Training about 1\% of the parameters on 12.9K questions, it outperforms all baseline training objectives in domain on three backbones and, on InternVL3-8B and Qwen3-VL-8B, surpasses all 10 general and medical VLMs on OOD open-ended questions. These results suggest that supervising the gold answer after the model's own reasoning is an effective way to combine SFT and RL.

\section*{Limitations}

\paragraph{Reasoning-answer consistency.} Supervising the gold answer after the model's own reasoning occasionally teaches answers that are inconsistent with that reasoning, and trained models sometimes report findings on normal OOD images, which are scarce in the training data. Gating supervision on consistency and adding more normal cases to training are possible directions for future work.

\paragraph{Evaluation.} Exact match often misses synonyms and abbreviations. A blinded two-physician audit with images and questions supports the LLM judge's decisions; EM and judged accuracy are complementary estimates, but neither fully replaces clinical evaluation.
Baselines differ in prompt formats and training corpora, so we report results with both a common prompt and each model's own prompt. On the external sets, rankings shift slightly with the prompt (\S\ref{sec:res-vlm}).

\paragraph{Data and scope.} The training set covers four sources but contains few normal cases, and is restricted to answers of at most three words. Fine-grained diagnosis remains hard: on VQA-Med 2020 every model scores below 13\% judged (\S\ref{sec:res-vlm}).

\section*{Ethics Statement}

All datasets are public research benchmarks with de-identified images, used under their licenses; no patient data was collected for this work, and only question and answer text, never images, was sent to an external LLM service. The models are research artifacts and must not be used for diagnosis or treatment without regulatory validation and expert oversight.

\bibliography{references}

\begin{thebibliography}{42}
\providecommand{\natexlab}[1]{#1}

\bibitem[{Alsentzer et~al.(2019)Alsentzer, Murphy, Boag, Weng, Jin, Naumann,
  and McDermott}]{clinicalbert_acl}
Emily Alsentzer, John Murphy, William Boag, Wei-Hung Weng, Di~Jin, Tristan
  Naumann, and Matthew McDermott. 2019.
\newblock \href {https://doi.org/10.18653/v1/W19-1909} {Publicly available
  clinical {BERT} embeddings}.
\newblock In \emph{Proceedings of the 2nd Clinical Natural Language Processing
  Workshop}, pages 72--78, Minneapolis, Minnesota, USA. Association for
  Computational Linguistics.

\bibitem[{Bai et~al.(2025)Bai, Cai, Chen, Chen, Chen, Cheng, Deng, Ding, Gao,
  Ge et~al.}]{qwen3vl}
Shuai Bai, Yuxuan Cai, Ruizhe Chen, Keqin Chen, Xionghui Chen, Zesen Cheng,
  Lianghao Deng, Wei Ding, Chang Gao, Chunjiang Ge, et~al. 2025.
\newblock \href {https://arxiv.org/abs/2511.21631} {{Qwen3-VL} technical
  report}.
\newblock \emph{arXiv preprint arXiv:2511.21631}.

\bibitem[{Ben~Abacha et~al.(2020)Ben~Abacha, Datla, Hasan, Demner-Fushman, and
  M{\"u}ller}]{vqamed2020}
Asma Ben~Abacha, Vivek~V. Datla, Sadid~A. Hasan, Dina Demner-Fushman, and
  Henning M{\"u}ller. 2020.
\newblock \href {https://ceur-ws.org/Vol-2696/paper_106.pdf} {Overview of the
  {VQA-Med} task at {ImageCLEF} 2020: Visual question answering and generation
  in the medical domain}.
\newblock In \emph{CLEF 2020 Working Notes}, volume 2696 of \emph{CEUR Workshop
  Proceedings}.

\bibitem[{Chen et~al.(2024{\natexlab{a}})Chen, Gui, Ouyang, Gao, Chen, Chen,
  Wang, Cai, Ji, Wan, and Wang}]{huatuogptvision}
Junying Chen, Chi Gui, Ruyi Ouyang, Anningzhe Gao, Shunian Chen, Guiming~Hardy
  Chen, Xidong Wang, Zhenyang Cai, Ke~Ji, Xiang Wan, and Benyou Wang.
  2024{\natexlab{a}}.
\newblock \href {https://doi.org/10.18653/v1/2024.emnlp-main.418} {Towards
  injecting medical visual knowledge into multimodal {LLMs} at scale}.
\newblock In \emph{Proceedings of the 2024 Conference on Empirical Methods in
  Natural Language Processing}, pages 7346--7370.

\bibitem[{Chen et~al.(2024{\natexlab{b}})Chen, Ye, Wang, Li, Deng, Li, Li,
  Duan, Huang, Su et~al.}]{gmaimmbench}
Pengcheng Chen, Jin Ye, Guoan Wang, Yanjun Li, Zhongying Deng, Wei Li, Tianbin
  Li, Haodong Duan, Ziyan Huang, Yanzhou Su, et~al. 2024{\natexlab{b}}.
\newblock \href {https://arxiv.org/abs/2408.03361} {{GMAI-MMBench:} a
  comprehensive multimodal evaluation benchmark towards general medical {AI}}.
\newblock In \emph{Advances in Neural Information Processing Systems (Datasets
  and Benchmarks Track)}, volume~37, pages 94327--94427.

\bibitem[{{DeepSeek-AI}(2026)}]{deepseekv4}
{DeepSeek-AI}. 2026.
\newblock \href {https://arxiv.org/abs/2606.19348} {{DeepSeek-V4}: Towards
  highly efficient million-token context intelligence}.
\newblock \emph{arXiv preprint arXiv:2606.19348}.

\bibitem[{Deria et~al.(2026)Deria, Kumar, Dukre, Segal, Khan, and
  Razzak}]{medmo}
Ankan Deria, Komal Kumar, Adinath~Madhavrao Dukre, Eran Segal, Salman Khan, and
  Imran Razzak. 2026.
\newblock \href {https://arxiv.org/abs/2602.06965} {{MedMO:} grounding and
  understanding multimodal large language model for medical images}.
\newblock In \emph{Proceedings of the IEEE/CVF Conference on Computer Vision
  and Pattern Recognition (CVPR)}, pages 5092--5103.

\bibitem[{Fu et~al.(2026)Fu, Chen, Chai, Wang, Tu, Yin, Lin, Zhang, Zhu, and
  Zhao}]{fu2025srft}
Yuqian Fu, Tinghong Chen, Jiajun Chai, Xihuai Wang, Songjun Tu, Guojun Yin, Wei
  Lin, Qichao Zhang, Yuanheng Zhu, and Dongbin Zhao. 2026.
\newblock \href {https://arxiv.org/abs/2506.19767} {{SRFT:} a single-stage
  method with supervised and reinforcement fine-tuning for reasoning}.
\newblock In \emph{International Conference on Learning Representations}.

\bibitem[{Gautam et~al.(2024)Gautam, Stor{\aa}s, Midoglu, Hicks, Thambawita,
  Halvorsen, and Riegler}]{kvasirvqa}
Sushant Gautam, Andrea~M. Stor{\aa}s, Cise Midoglu, Steven~A. Hicks, Vajira
  Thambawita, P{\aa}l Halvorsen, and Michael~A. Riegler. 2024.
\newblock \href {https://doi.org/10.1145/3689096.3689458} {{Kvasir-VQA:} a
  text-image pair {GI} tract dataset}.
\newblock In \emph{Proceedings of the First International Workshop on
  Vision-Language Models for Biomedical Applications (VLM4Bio)}, pages 3--12.
  ACM.

\bibitem[{{Gemma Team} et~al.(2025){Gemma Team}, Kamath, Ferret, Pathak,
  Vieillard, Merhej, Perrin, Matejovicova, Ram{\'e}, Rivi{\`e}re
  et~al.}]{gemma3}
{Gemma Team}, Aishwarya Kamath, Johan Ferret, Shreya Pathak, Nino Vieillard,
  Ramona Merhej, Sarah Perrin, Tatiana Matejovicova, Alexandre Ram{\'e},
  Morgane Rivi{\`e}re, et~al. 2025.
\newblock \href {https://arxiv.org/abs/2503.19786} {{Gemma} {3} technical
  report}.
\newblock \emph{arXiv preprint arXiv:2503.19786}.

\bibitem[{Guo et~al.(2025)Guo, Yang, Zhang, Song, Wang, Zhu, Xu, Zhang, Ma, Bi
  et~al.}]{guo2025deepseekr1}
Daya Guo, Dejian Yang, Haowei Zhang, Junxiao Song, Peiyi Wang, Qihao Zhu,
  Runxin Xu, Ruoyu Zhang, Shirong Ma, Xiao Bi, et~al. 2025.
\newblock \href {https://doi.org/10.1038/s41586-025-09422-z} {{DeepSeek-R1}
  incentivizes reasoning in {LLMs} through reinforcement learning}.
\newblock \emph{Nature}, 645(8081):633--638.

\bibitem[{Han et~al.(2023)Han, Han, and {Unsloth team}}]{unsloth}
Daniel Han, Michael Han, and {Unsloth team}. 2023.
\newblock Unsloth.
\newblock \url{https://github.com/unslothai/unsloth}.

\bibitem[{Han et~al.(2026)Han, Zhou, Gao, Zhou, Li, Kumar, Fan, Li, and
  Zhang}]{ebpo}
Kevin Han, Yuhang Zhou, Mingze Gao, Gedi Zhou, Serena Li, Abhishek Kumar,
  Xiangjun Fan, Weiwei Li, and Lizhu Zhang. 2026.
\newblock \href {https://arxiv.org/abs/2602.05165} {{EBPO:} empirical {Bayes}
  shrinkage for stabilizing group-relative policy optimization}.
\newblock \emph{arXiv preprint arXiv:2602.05165}.

\bibitem[{He et~al.(2020)He, Zhang, Mou, Xing, and Xie}]{pathvqa}
Xuehai He, Yichen Zhang, Luntian Mou, Eric Xing, and Pengtao Xie. 2020.
\newblock \href {https://arxiv.org/abs/2003.10286} {{PathVQA:} {30000+}
  questions for medical visual question answering}.
\newblock \emph{arXiv preprint arXiv:2003.10286}.

\bibitem[{Hu et~al.(2022)Hu, Shen, Wallis, Allen-Zhu, Li, Wang, Wang, and
  Chen}]{hu2022lora}
Edward~J. Hu, Yelong Shen, Phillip Wallis, Zeyuan Allen-Zhu, Yuanzhi Li, Shean
  Wang, Lu~Wang, and Weizhu Chen. 2022.
\newblock \href {https://arxiv.org/abs/2106.09685} {{LoRA:} low-rank adaptation
  of large language models}.
\newblock In \emph{International Conference on Learning Representations}.

\bibitem[{Hu et~al.(2024)Hu, Li, Lu, Shao, He, Qiao, and Luo}]{omnimedvqa}
Yutao Hu, Tianbin Li, Quanfeng Lu, Wenqi Shao, Junjun He, Yu~Qiao, and Ping
  Luo. 2024.
\newblock \href {https://doi.org/10.1109/CVPR52733.2024.02093} {{OmniMedVQA:} a
  new large-scale comprehensive evaluation benchmark for medical {LVLM}}.
\newblock In \emph{Proceedings of the IEEE/CVF Conference on Computer Vision
  and Pattern Recognition (CVPR)}, pages 22170--22183.

\bibitem[{Huang et~al.(2026)Huang, Wu, Liu, Tang, and Zhou}]{medvlthinker}
Xiaoke Huang, Juncheng Wu, Hui Liu, Xianfeng Tang, and Yuyin Zhou. 2026.
\newblock \href {https://proceedings.mlr.press/v297/huang26b.html}
  {{MedVLThinker:} simple baselines for multimodal medical reasoning}.
\newblock In \emph{Proceedings of the Fifth Machine Learning for Health
  Symposium}, volume 297 of \emph{Proceedings of Machine Learning Research},
  pages 384--398.

\bibitem[{Lai et~al.(2026)Lai, Zhong, Li, Zhao, Li, Psounis, and Yang}]{medr1}
Yuxiang Lai, Jike Zhong, Ming Li, Shitian Zhao, Yuheng Li, Konstantinos
  Psounis, and Xiaofeng Yang. 2026.
\newblock \href {https://doi.org/10.1109/TMI.2026.3661001} {{Med-R1:}
  reinforcement learning for generalizable medical reasoning in vision-language
  models}.
\newblock \emph{IEEE Transactions on Medical Imaging}, 45(6):2727--2737.

\bibitem[{Lau et~al.(2018)Lau, Gayen, Ben~Abacha, and Demner-Fushman}]{vqarad}
Jason~J. Lau, Soumya Gayen, Asma Ben~Abacha, and Dina Demner-Fushman. 2018.
\newblock \href {https://doi.org/10.1038/sdata.2018.251} {A dataset of
  clinically generated visual questions and answers about radiology images}.
\newblock \emph{Scientific Data}, 5(1):180251.

\bibitem[{Le et~al.(2026)Le, Jeon, Vu, Lai, and Yang}]{rlzvp}
Thanh-Long~V. Le, Myeongho Jeon, Kim Vu, Viet Lai, and Eunho Yang. 2026.
\newblock \href {https://arxiv.org/abs/2509.21880} {No prompt left behind:
  Exploiting zero-variance prompts in {LLM} reinforcement learning via
  entropy-guided advantage shaping}.
\newblock In \emph{International Conference on Learning Representations}.

\bibitem[{Liu et~al.(2021)Liu, Zhan, Xu, Ma, Yang, and Wu}]{slake}
Bo~Liu, Li-Ming Zhan, Li~Xu, Lin Ma, Yan Yang, and Xiao-Ming Wu. 2021.
\newblock \href {https://doi.org/10.1109/ISBI48211.2021.9434010} {{SLAKE:} a
  semantically-labeled knowledge-enhanced dataset for medical visual question
  answering}.
\newblock In \emph{IEEE 18th International Symposium on Biomedical Imaging
  (ISBI)}, pages 1650--1654.

\bibitem[{Liu et~al.(2024)Liu, Li, Li, and Lee}]{llava15}
Haotian Liu, Chunyuan Li, Yuheng Li, and Yong~Jae Lee. 2024.
\newblock \href {https://doi.org/10.1109/CVPR52733.2024.02484} {Improved
  baselines with visual instruction tuning}.
\newblock In \emph{Proceedings of the IEEE/CVF Conference on Computer Vision
  and Pattern Recognition (CVPR)}, pages 26296--26306.

\bibitem[{Liu et~al.(2026)Liu, Chen, Liu, Zhu, Zhong, Yu, and Jia}]{visurf}
Yuqi Liu, Liangyu Chen, Jiazhen Liu, Mingkang Zhu, Zhisheng Zhong, Bei Yu, and
  Jiaya Jia. 2026.
\newblock \href {https://proceedings.mlr.press/v306/liu26p.html} {{ViSurf:}
  visual supervised-and-reinforcement fine-tuning for large vision-and-language
  models}.
\newblock In \emph{Proceedings of the 43rd International Conference on Machine
  Learning}, volume 306 of \emph{Proceedings of Machine Learning Research},
  pages 75698--75710.

\bibitem[{Lv et~al.(2025)Lv, Zuo, Sun, Liu, Wei, Chen, Zhu, Zhang, Wang, Ding,
  and Zhou}]{lv2025hpt}
Xingtai Lv, Yuxin Zuo, Youbang Sun, Hongyi Liu, Yuntian Wei, Zhekai Chen,
  Xuekai Zhu, Kaiyan Zhang, Bingning Wang, Ning Ding, and Bowen Zhou. 2025.
\newblock \href {https://arxiv.org/abs/2509.04419} {Towards a unified view of
  large language model post-training}.
\newblock \emph{arXiv preprint arXiv:2509.04419}.

\bibitem[{Mullappilly et~al.(2026)Mullappilly, Kurpath, Mohamed, Zidan, Khan,
  Khan, Anwer, and Cholakkal}]{medixr1}
Sahal~Shaji Mullappilly, Mohammed~Irfan Kurpath, Omair Mohamed, Mohamed Zidan,
  Fahad Khan, Salman Khan, Rao Anwer, and Hisham Cholakkal. 2026.
\newblock \href {https://arxiv.org/abs/2602.23363} {{MediX-R1:} open ended
  medical reinforcement learning}.
\newblock In \emph{Findings of the Association for Computational Linguistics:
  EMNLP 2026}.
\newblock To appear.

\bibitem[{Pan et~al.(2025)Pan, Liu, Wu, Liu, Zhu, Li, Chen, Ouyang, and
  Rueckert}]{medvlmr1}
Jiazhen Pan, Che Liu, Junde Wu, Fenglin Liu, Jiayuan Zhu, Hongwei~Bran Li, Chen
  Chen, Cheng Ouyang, and Daniel Rueckert. 2025.
\newblock \href {https://doi.org/10.1007/978-3-032-04981-0_32} {{MedVLM-R1:}
  incentivizing medical reasoning capability of vision-language models {(VLMs)}
  via reinforcement learning}.
\newblock In \emph{Medical Image Computing and Computer Assisted Intervention
  (MICCAI)}, volume 15966 of \emph{Lecture Notes in Computer Science}.

\bibitem[{Rui et~al.(2025)Rui, Chen, Ma, and Wang}]{medcco}
Shaohao Rui, Kaitao Chen, Weijie Ma, and Xiaosong Wang. 2025.
\newblock \href {https://arxiv.org/abs/2505.19213} {Improving medical reasoning
  with curriculum-aware reinforcement learning}.
\newblock \emph{arXiv preprint arXiv:2505.19213}.

\bibitem[{Sellergren et~al.(2025)Sellergren, Kazemzadeh, Jaroensri, Kiraly,
  Traverse, Kohlberger, Xu, Jamil, Hughes, Lau et~al.}]{medgemma}
Andrew Sellergren, Sahar Kazemzadeh, Tiam Jaroensri, Atilla Kiraly, Madeleine
  Traverse, Timo Kohlberger, Shawn Xu, Fayaz Jamil, C{\'i}an Hughes, Charles
  Lau, et~al. 2025.
\newblock \href {https://arxiv.org/abs/2507.05201} {{MedGemma} technical
  report}.
\newblock \emph{arXiv preprint arXiv:2507.05201}.

\bibitem[{Shao et~al.(2024)Shao, Wang, Zhu, Xu, Song, Bi, Zhang, Zhang, Li, Wu,
  and Guo}]{shao2024deepseekmath}
Zhihong Shao, Peiyi Wang, Qihao Zhu, Runxin Xu, Junxiao Song, Xiao Bi, Haowei
  Zhang, Mingchuan Zhang, Y.~K. Li, Y.~Wu, and Daya Guo. 2024.
\newblock \href {https://arxiv.org/abs/2402.03300} {{DeepSeekMath:} pushing the
  limits of mathematical reasoning in open language models}.
\newblock \emph{arXiv preprint arXiv:2402.03300}.

\bibitem[{Shu et~al.(2025)Shu, Liu, Chen, Li, and Dai}]{flemingvl}
Yan Shu, Chi Liu, Robin Chen, Derek Li, and Bryan Dai. 2025.
\newblock \href {https://arxiv.org/abs/2511.00916} {{Fleming-VL:} towards
  universal medical visual reasoning with multimodal {LLMs}}.
\newblock \emph{arXiv preprint arXiv:2511.00916}.

\bibitem[{Tang et~al.(2025)Tang, Wang, Madaan, and Munos}]{tang2025jepo}
Yunhao Tang, Sid Wang, Lovish Madaan, and R{\'e}mi Munos. 2025.
\newblock \href {https://arxiv.org/abs/2503.19618} {Beyond verifiable rewards:
  Scaling reinforcement learning in language models to unverifiable data}.
\newblock In \emph{Advances in Neural Information Processing Systems}.

\bibitem[{von Werra et~al.(2020)von Werra, Belkada, Tunstall, Beeching, Thrush,
  Lambert, Huang, Rasul, and Gallou{\'e}dec}]{trl}
Leandro von Werra, Younes Belkada, Lewis Tunstall, Edward Beeching, Tristan
  Thrush, Nathan Lambert, Shengyi Huang, Kashif Rasul, and Quentin
  Gallou{\'e}dec. 2020.
\newblock {TRL}: Transformers reinforcement learning.
\newblock \url{https://github.com/huggingface/trl}.

\bibitem[{Xu et~al.(2026)Xu, Chan, Li, Aljunied, Yuan, Wang, Xiao, Chen, Liu,
  Li et~al.}]{lingshu}
Weiwen Xu, Hou~Pong Chan, Long Li, Mahani Aljunied, Ruifeng Yuan, Jianyu Wang,
  Chenghao Xiao, Guizhen Chen, Chaoqun Liu, Zhaodonghui Li, et~al. 2026.
\newblock \href {https://arxiv.org/abs/2506.07044} {{Lingshu:} a generalist
  foundation model for unified multimodal medical understanding and reasoning}.
\newblock \emph{IEEE Transactions on Pattern Analysis and Machine
  Intelligence}.
\newblock To appear.

\bibitem[{Yan et~al.(2025)Yan, Li, Hu, Wang, Cui, Qu, Cheng, and
  Zhang}]{yan2025luffy}
Jianhao Yan, Yafu Li, Zican Hu, Zhi Wang, Ganqu Cui, Xiaoye Qu, Yu~Cheng, and
  Yue Zhang. 2025.
\newblock \href {https://arxiv.org/abs/2504.14945} {Learning to reason under
  off-policy guidance}.
\newblock In \emph{Advances in Neural Information Processing Systems}.

\bibitem[{Yu et~al.(2025)Yu, Zhang, Zhu, Yuan, Zuo, Yue, Dai, Fan, Liu, Liu
  et~al.}]{yu2025dapo}
Qiying Yu, Zheng Zhang, Ruofei Zhu, Yufeng Yuan, Xiaochen Zuo, Yu~Yue, Weinan
  Dai, Tiantian Fan, Gaohong Liu, Juncai Liu, et~al. 2025.
\newblock \href {https://arxiv.org/abs/2503.14476} {{DAPO:} an open-source
  {LLM} reinforcement learning system at scale}.
\newblock In \emph{Advances in Neural Information Processing Systems}.

\bibitem[{Yuan et~al.(2023)Yuan, Yuan, Li, Dong, Lu, Tan, Zhou, and
  Zhou}]{yuan2023rft}
Zheng Yuan, Hongyi Yuan, Chengpeng Li, Guanting Dong, Keming Lu, Chuanqi Tan,
  Chang Zhou, and Jingren Zhou. 2023.
\newblock \href {https://arxiv.org/abs/2308.01825} {Scaling relationship on
  learning mathematical reasoning with large language models}.
\newblock \emph{arXiv preprint arXiv:2308.01825}.

\bibitem[{Zhang et~al.(2025)Zhang, Xu, Usuyama, Xu, Bagga, Tinn, Preston, Rao,
  Wei, Valluri et~al.}]{biomedclip}
Sheng Zhang, Yanbo Xu, Naoto Usuyama, Hanwen Xu, Jaspreet Bagga, Robert Tinn,
  Sam Preston, Rajesh Rao, Mu~Wei, Naveen Valluri, et~al. 2025.
\newblock \href {https://doi.org/10.1056/AIoa2400640} {A multimodal biomedical
  foundation model trained from fifteen million image--text pairs}.
\newblock \emph{NEJM AI}, 2(1).

\bibitem[{Zhang et~al.(2020)Zhang, Kishore, Wu, Weinberger, and
  Artzi}]{bertscore}
Tianyi Zhang, Varsha Kishore, Felix Wu, Kilian~Q. Weinberger, and Yoav Artzi.
  2020.
\newblock \href {https://arxiv.org/abs/1904.09675} {{BERTScore}: Evaluating
  text generation with {BERT}}.
\newblock In \emph{International Conference on Learning Representations}.

\bibitem[{Zheng et~al.(2025)Zheng, Liu, Li, Chen, Yu, Gao, Dang, Liu, Men,
  Yang, Zhou, and Lin}]{zheng2025gspo}
Chujie Zheng, Shixuan Liu, Mingze Li, Xiong-Hui Chen, Bowen Yu, Chang Gao, Kai
  Dang, Yuqiong Liu, Rui Men, An~Yang, Jingren Zhou, and Junyang Lin. 2025.
\newblock \href {https://arxiv.org/abs/2507.18071} {Group sequence policy
  optimization}.
\newblock \emph{arXiv preprint arXiv:2507.18071}.

\bibitem[{Zhou et~al.(2026)Zhou, Liu, Sims, Wang, Pang, Li, Wang, Lin, and
  Du}]{zhou2025verifree}
Xiangxin Zhou, Zichen Liu, Anya Sims, Haonan Wang, Tianyu Pang, Chongxuan Li,
  Liang Wang, Min Lin, and Chao Du. 2026.
\newblock \href {https://arxiv.org/abs/2505.21493} {Reinforcing general
  reasoning without verifiers}.
\newblock In \emph{International Conference on Learning Representations}.

\bibitem[{Zhu et~al.(2025)Zhu, Wang, Chen, Liu, Ye, Gu, Tian, Duan, Su, Shao
  et~al.}]{internvl3}
Jinguo Zhu, Weiyun Wang, Zhe Chen, Zhaoyang Liu, Shenglong Ye, Lixin Gu, Hao
  Tian, Yuchen Duan, Weijie Su, Jie Shao, et~al. 2025.
\newblock \href {https://arxiv.org/abs/2504.10479} {{InternVL3:} exploring
  advanced training and test-time recipes for open-source multimodal models}.
\newblock \emph{arXiv preprint arXiv:2504.10479}.

\bibitem[{{ZJU4HealthCare}(2026)}]{healthgptpro}
{ZJU4HealthCare}. 2026.
\newblock {HealthGPT-Pro:} a high-performance multimodal large language model
  for medical understanding and analysis.
\newblock
  \url{https://github.com/ZJU4HealthCare/HealthGPT/tree/main/HealthGPT-Pro}.

\end{thebibliography}

\appendix
\makeatletter
\newcommand{\grasp@app@section}{\appendixname~}
\renewcommand{\@seccntformat}[1]{%
  \ifcsname grasp@app@#1\endcsname\csname grasp@app@#1\endcsname\fi\csname the#1\endcsname\quad}
\makeatother

\section{Method and Training Details}
\label{app:method}
\label{app:prompts}
\label{app:hparams}

Table~\ref{tab:hparams} lists the hyperparameters of the main run.

\begin{table}[t]
\centering
\footnotesize
\setlength{\tabcolsep}{3pt}
\begin{tabularx}{\columnwidth}{@{}>{\raggedright\arraybackslash}p{2.5cm}>{\raggedright\arraybackslash}X@{}}
\toprule
Backbone / adapters & Qwen3-VL-8B-Instruct; BF16 LoRA $r{=}32$, $\alpha{=}32$ on attention and MLP layers; 102.7M parameters \\
Sequence / generation & 4,096 / 256 tokens \\
Group / batch & 8 rollouts; 12 questions per step \\
Sampling & temperature 1.15; fresh rollouts each step \\
Optimizer & 8-bit AdamW \\
Learning rate & $2{\times}10^{-5}$, cosine, 10\% warm-up \\
Epochs / steps & 2 / 2,148 \\
KL / clipping & $\beta{=}0.02$; $\epsilon_l{=}3{\times}10^{-4}$, $\epsilon_h{=}4{\times}10^{-4}$ \\
$\lambda_{\max},\gamma,\kappa,\eta$ & 0.4, 1, 1, 0.99 \\
Hardware / time & 4$\times$A100 80GB; ${\approx}$20 h (${\approx}$81 GPU-h) \\
\bottomrule
\end{tabularx}
\caption{\textbf{Hyperparameters of the main run.} Our implementation builds on TRL~\citep{trl} and Unsloth~\citep{unsloth}.}
\label{tab:hparams}
\end{table}

\paragraph{Loss aggregation.} We average the policy-gradient term over all rollouts in a batch. The weighted answer losses are summed over groups and divided by the total number of supervised answer tokens. Only the policy-gradient term is divided by $s_{\text{RL}}$, an EMA of the advantage RMS. $s_{\text{SFT}}$ is the mean answer NLL over the first 20 steps and stays fixed afterwards. We normalize rewards at the batch level so that groups with nearly identical rewards are not stretched to unit variance.

\paragraph{Reasoning prefixes.} The supervised loss masks the reasoning tokens and their tags. When a rollout's reasoning is unusable, we truncate it if it loops and otherwise borrow a valid reasoning from the same group, the same batch or a bank of the 256 most recent traces. In the main run, 98.1\% of the supervised rollouts keep their own reasoning unchanged, 1.9\% use a truncated version and 0.01\% use a borrowed one.

\paragraph{Answer supervision as a bound.} By Jensen's inequality, $\log\pi_\theta(a^\star\mid x)\ge\mathbb{E}_{z\sim\pi_\theta}[\log\pi_\theta(a^\star\mid x,z)]$. For a fixed reasoning distribution, the SFT branch increases the right-hand side and thereby this lower bound. Verifier-free RL methods use the likelihood of the reference answer after sampled reasoning in place of a verifier, either through this bound~\citep{tang2025jepo} or through the probability itself~\citep{zhou2025verifree}. \method instead combines the bound with verifiable rewards.

\paragraph{Extreme-group correction.} Both $\mu_t$ and $s_t$ in Eq.~\ref{eq:zvc} are exponential moving averages updated at every step:
\begin{equation}
\begin{aligned}
    \mu_t&=\eta\,\mu_{t-1}+(1-\eta)\,\bar p,\\
    s_t&=\eta\,s_{t-1}+(1-\eta)\sqrt{\tfrac{1}{N}\textstyle\sum_{j=1}^{N}(A^{\text{rel}}_j)^2},
\end{aligned}
    \label{eq:ema}
\end{equation}
where $\bar p$ is the mean group accuracy of the current batch, $N$ the number of rollouts in the batch and $\eta{=}0.99$. Averaged over training batches, the correction is 0.0002, compared with a mean absolute value of 0.016 per group, so in practice it is nearly zero-mean.

\paragraph{Prompt and scoring.} Our models (in training and evaluation), the reasoning baselines and all models under the shared protocol receive a single user turn, the image followed by

\begin{quote}\small\ttfamily
\{question\}. Also first provide your reasoning in 8 to 40 words on how you solve the question between <REASONING> and </REASONING> and then your final answer between <answer> and (put a single yes or no for close-ended question and one to three words for open-ended questions) </answer>
\end{quote}

\noindent The prompt asks for a reasoning $z$ of 8--40 words, and no system prompt is used. Under the own-prompt protocol, direct-answer baselines instead receive the question followed by ``Answer the question using a single word or phrase,'' and we score the first line of their response. All models decode greedily, and for our models the answer is read from the \texttt{<answer>} tags. Before exact matching, we lower-case both strings, map the number words one to ten to digits, remove ASCII punctuation and collapse whitespace.

\paragraph{Rewards.} All runs use the reward of Eq.~\ref{eq:reward}. $r_{\text{fmt}}$ checks that the reasoning and answer tags each appear once, that the reasoning and the answer have the required lengths and that they come in the right order, and it penalizes loops, truncation and empty reasoning. $r_{\text{EM}}$ is 2 for a normalized exact match and 0 otherwise. $r_{\text{syn}}$ is the Bio\_ClinicalBERT BERTScore F1, clipped at zero, and $r_{\text{img}}$ is the BiomedCLIP cosine similarity between the image and the reasoning, clipped to $[0,1]$. Ablations set the weight of the removed term to zero.

\paragraph{Training seeds.} We train the main configuration, SFT and GSPO with three seeds each, keeping data, rewards and optimization settings fixed. Table~\ref{tab:app-seeds} reports the mean and standard deviation over the three runs, and the main tables report one of these runs.

\begin{table}[t]
\centering
\footnotesize
\setlength{\tabcolsep}{3.2pt}
\newcommand{\pmsd}[1]{{\scriptsize$\,\pm\,$#1}}
\begin{tabular}{@{}l ccc@{}}
\toprule
\textbf{Run} & L-all & L-open & O-open \\
\midrule
\rowcolor{oursrow}
\method ($n{=}3$) & 71.35\pmsd{0.56} & 39.72\pmsd{0.27} & 25.34\pmsd{0.24} \\
SFT ($n{=}3$) & 62.89\pmsd{0.31} & 33.35\pmsd{0.46} & 22.91\pmsd{0.06} \\
GSPO ($n{=}3$) & 59.80\pmsd{0.67} & 25.36\pmsd{1.04} & 22.70\pmsd{0.40} \\
\bottomrule
\end{tabular}
\caption{\textbf{Seed variation} (EM, \%; mean $\pm$ standard deviation over three training seeds). L: in domain; O: OmniMedVQA; open: open-ended questions.}
\label{tab:app-seeds}
\label{tab:app-seeds-base}
\end{table}

\section{Data Construction}
\label{app:data}

\paragraph{Training set.} \dataname draws on the official training splits of PathVQA (4,430 retained questions), SLAKE (4,184 English questions) and VQA-RAD (1,452), plus 2,827 items from the GMAI-MMBench validation set. We keep questions that can be answered without options and whose answers have one to three words. DeepSeek-V4 rewrites option-dependent questions as direct questions. It sees only the question, options and answer, never the image, and we discard rewrites that remain ambiguous.

\paragraph{Evaluation splits.} The in-domain test data consist of a held-out validation set of 1,389 questions and the SLAKE, VQA-RAD and PathVQA test splits with 991, 413 and 3,807 questions (6,600 in total, 2,056 of them open-ended). For the OOD set, we convert the open-access part of OmniMedVQA, which covers eight imaging modalities, with the same short-answer rule. In 1,033 questions the gold answer was copied from an option label and does not answer the question as worded, so we correct it: 781 normal yes/no cases become \emph{no}, 46 abnormal yes/no cases become \emph{yes}, 24 normal/abnormal questions receive these literal labels, and 182 open questions with non-specific labels are removed. The final OOD set has 11,575 questions, 85\% of them open-ended. Since 93\% of its 1,765 yes/no questions have the answer \emph{no}, we report balanced accuracy on them. Here OOD means a held-out source with its own question distribution, not an unseen imaging modality. The external sets are the Kvasir-VQA test set (1,346 questions on 183 endoscopy images, none of them used in training) and 353 diagnosis questions from VQA-Med 2020, after removing six images that also appear in training.

\paragraph{Release.} We release the 12,893 training items and the converted OmniMedVQA questions with their processing type, source, original question and image identifiers and corrected answers at \dataurl. The images remain under the licenses of the original datasets and are not redistributed.

\section{Additional Results}
\label{app:results}

\paragraph{Own-prompt comparison.} Table~\ref{tab:app-vlm-own} repeats the comparison of Table~\ref{tab:vlm} with each model's own prompt. \method is still the best on OOD open-ended questions and on Kvasir-VQA. In domain, only HealthGPT-Pro scores higher (73.3\% vs.\ 71.9\% EM on all questions), but it trains over 80 times as many parameters (8.8B vs.\ 0.1B) on about 1,000 times as much data (13M vs.\ 12.9K examples), and it trails \method on OOD open-ended questions (\S\ref{sec:res-vlm}).

\begin{table*}[t]
\centering
\footnotesize
\setlength{\tabcolsep}{3.1pt}
\begin{tabular}{@{}l rr cc cc cc cc cc@{}}
\toprule
& & & \multicolumn{4}{c}{\textbf{In domain}} & \multicolumn{4}{c}{\textbf{OmniMedVQA}} & \multicolumn{2}{c}{\textbf{Kvasir}} \\
\cmidrule(lr){4-7}\cmidrule(lr){8-11}\cmidrule(l){12-13}
& & & \multicolumn{2}{c}{Open} & \multicolumn{2}{c}{All} & \multicolumn{2}{c}{Open} & \multicolumn{2}{c}{All} & \multicolumn{2}{c}{All} \\
\cmidrule(lr){4-5}\cmidrule(lr){6-7}\cmidrule(lr){8-9}\cmidrule(lr){10-11}\cmidrule(l){12-13}
\textbf{Model} & \textbf{Tuned} & \textbf{Data} & EM & LLM & EM & LLM & EM & LLM & EM & LLM & EM & LLM \\
\midrule
Qwen3-VL-8B       & --    & --     & 26.8 & 33.1 & 55.7 & 57.6 & 20.4 & 28.5 & 22.3 & 29.2 & 44.9 & 45.7 \\
InternVL3-8B      & --    & --     & 32.0 & 35.5 & 56.6 & 57.7 & 21.3 & 28.1 & 25.3 & 31.1 & 40.3 & 41.0 \\
MedGemma-4B       & 4.3B  & 33.6M  & 26.8 & 35.6 & 54.0 & 56.8 & 15.8 & 26.9 & 23.2 & 32.6 & 19.1 & 19.2 \\
MedGemma-27B      & ${\approx}$27B & -- & 26.5 & 34.8 & 53.9 & 56.5 & 18.2 & 25.2 & 17.2 & 23.2 & 27.3 & 30.9 \\
HuatuoGPT-V-7B    & 8.3B  & 1.3M   & 24.0 & 30.8 & 53.9 & 56.0 & 16.1 & 24.8 & 26.2 & 33.5 & 33.6 & 34.0 \\
HuatuoGPT-V-34B   & ${\approx}$34B & ${>}$1.6M & 22.8 & 29.2 & 55.5 & 57.5 & 17.6 & 24.8 & 28.7 & 34.8 & 44.9 & 45.4 \\
Lingshu-7B        & 8.3B  & 12.1M  & 38.0 & 42.2 & 68.6 & 69.8 & 17.8 & 26.4 & 26.4 & 33.7 & 41.5 & 41.9 \\
HealthGPT-Pro-8B  & 8.8B  & 13M    & 39.6 & \textbf{44.8} & \textbf{73.3} & \textbf{74.9} & 21.6 & 28.3 & 30.9 & 36.5 & 38.7 & 39.2 \\
\rowcolor{oursrow}
\textbf{\method-Qwen (Ours)} & 0.1B  & 12.9K  & 40.0 & 42.7 & 71.9 & 72.6 & 25.5 & 31.9 & 28.6 & 33.9 & \textbf{45.1} & \textbf{45.9} \\
\rowcolor{oursrow}
\textbf{\method-InternVL (Ours)} & 0.1B & 12.9K & \textbf{40.1} & 42.5 & 71.3 & 72.0 & \textbf{27.3} & \textbf{34.1} & \textbf{31.3} & \textbf{37.0} & 43.8 & 43.9 \\
\bottomrule
\end{tabular}
\caption{\textbf{Comparison with VLMs under each model's own prompt.} Direct-answer models are asked for a single word or phrase, while reasoning models and \method use their own reasoning format. Columns and \method rows as in Table~\ref{tab:vlm}; ${>}$: lower bound, as HuatuoGPT-V-34B also trains on medical text of unreported size.}
\label{tab:app-vlm-own}
\end{table*}

\begin{figure}[!htb]
    \centering
    \includegraphics[width=\columnwidth]{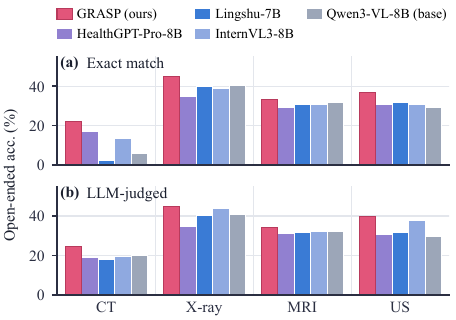}
    \caption{\textbf{OOD open-ended accuracy on the radiology modalities} (OmniMedVQA, own prompt): (a) exact match and (b) LLM-judged. \method: \method-Qwen.}
    \label{fig:oodmod}
\end{figure}

\paragraph{OOD accuracy by modality.} Figure~\ref{fig:oodmod} splits OOD open-ended accuracy by modality for the four radiology modalities. \method is the best on each of them, under exact match as well as under the LLM judge.

\begin{figure}[!htb]
    \centering
    \includegraphics[width=\columnwidth]{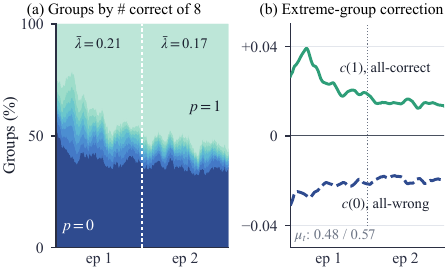}
    \caption{\textbf{A curriculum without a schedule} (main run, 60-step average). (a) Rollout groups by number of correct answers; $\bar\lambda$: supervision weight averaged over all groups in each epoch. (b) Training-adaptive extreme-group correction.}
    \label{fig:dynamics}
\end{figure}

\paragraph{Training dynamics.} Figure~\ref{fig:dynamics} shows how the share of each group type, the supervision weight and the extreme-group correction change during the main run (\S\ref{sec:dynamics}).

\begin{figure}[!htb]
    \centering
    \includegraphics[width=\columnwidth]{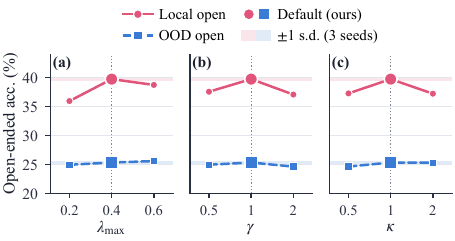}
    \caption{\textbf{Sensitivity to $\lambda_{\max}$, $\gamma$ and $\kappa$} (open-ended EM, two epochs). Each non-default point is a single run that changes only the indicated parameter. The default point is the three-seed mean, and the shaded bands show $\pm1$ standard deviation.}
    \label{fig:sensitivity}
\end{figure}

\paragraph{Hyperparameter sweep.} Figure~\ref{fig:sensitivity} and Table~\ref{tab:app-sweep} report the sweep over $\lambda_{\max}$, $\gamma$ and $\kappa$, changing one value at a time (\S\ref{sec:sensitivity}).

\begin{table}[!htb]
\centering
\footnotesize
\setlength{\tabcolsep}{3.2pt}
\begin{tabular}{@{}l ccc@{}}
\toprule
\textbf{Setting} & L-all & L-open & O-open \\
\midrule
\rowcolor{oursrow}
default ($\lambda_{\max}{=}0.4,\gamma{=}1,\kappa{=}1$) & 71.4 & 39.7 & 25.3 \\
$\lambda_{\max}{=}0.2$ & 68.7 & 35.9 & 25.0 \\
$\lambda_{\max}{=}0.6$ & 69.8 & 38.7 & 25.6 \\
$\gamma{=}0.5$ & 69.4 & 37.5 & 25.0 \\
$\gamma{=}2$ & 68.4 & 37.1 & 24.6 \\
$\kappa{=}0.5$ & 69.0 & 37.3 & 24.6 \\
$\kappa{=}2$ & 69.1 & 37.2 & 25.3 \\
\bottomrule
\end{tabular}
\caption{\textbf{Hyperparameter sweep} (EM, \%; two epochs). Each non-default row changes one value; the default row is the three-seed mean.}
\label{tab:app-sweep}
\end{table}

\section{Qualitative Examples}
\label{app:qualitative}

\begin{figure*}[t]
\centering
\footnotesize
\setlength{\tabcolsep}{4pt}
\renewcommand{\arraystretch}{1.1}
\newcommand{\ok}{\textcolor{oursblue}{\cmark}}
\newcommand{\no}{\textcolor{red!70!black}{\xmark}}
\newcommand{\qimg}[1]{\includegraphics[width=2.3cm,height=2.3cm,keepaspectratio]{figures/qualitative/#1}}
\begin{tabularx}{\textwidth}{@{}>{\centering\arraybackslash}m{2.4cm} >{\raggedright\arraybackslash}p{3.3cm} >{\raggedright\arraybackslash}X >{\raggedright\arraybackslash}p{3.4cm}@{}}
\toprule
\textbf{Image} & \textbf{Question / gold} & \textbf{\method: reasoning $\rightarrow$ answer} & \textbf{Other models} \\
\midrule
\qimg{} &
\textbf{SLAKE (in domain).} What disease is/are shown on the right of lung? \newline Gold: \emph{Pneumonia} &
``The X-ray shows increased opacity and consolidation in the right lower lung field, suggesting pneumonia or atelectasis. These findings indicate lung tissue inflammation or collapse, commonly seen in infectious or obstructive conditions.'' $\rightarrow$ \emph{Pneumonia} \ok &
SFT: \emph{lung opacity} \no \newline \rlname: \emph{Atelectasis} \no \newline Lingshu: \emph{Pleural Effusion} \no \\
\midrule
\qimg{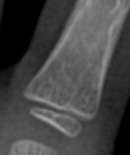} &
\textbf{OmniMedVQA X-ray (OOD).} From which joint does the bone in the image originate? \newline Gold: \emph{Proximal Interphalangeal} &
``The image shows a radiograph of a finger joint with a visible proximal phalanx and metacarpal. The joint appears to be the proximal interphalangeal (PIP) joint, located between the proximal and middle phalanges.'' $\rightarrow$ \emph{proximal interphalangeal} \ok &
SFT: \emph{metacarpophalangeal} \no \newline Lingshu, HealthGPT-Pro: \emph{Elbow} \no \\
\midrule
\qimg{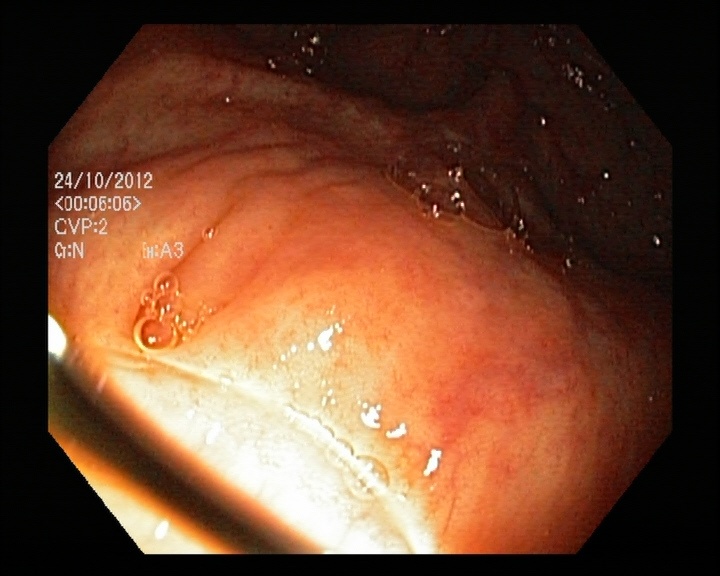} &
\textbf{Kvasir-VQA (held out).} What type of procedure is the image taken from? \newline Gold: \emph{colonoscopy} &
``The image shows a close-up view of internal mucosal tissue with medical metadata (date, time, CVP). This is characteristic of endoscopic procedures, likely colonoscopy or gastroscopy, used for internal organ examination.'' $\rightarrow$ \emph{colonoscopy} \ok &
Qwen3-VL base: \emph{Endoscopic procedure} \no \newline Fleming-VL: \emph{The image is taken from an endoscopic procedure.} \no \newline Lingshu: \emph{Upper Endoscopy} \no \\
\bottomrule
\end{tabularx}
\caption{\textbf{Qualitative examples} with verbatim greedy outputs. Marks follow exact match. Each example was selected because \method answers correctly and its reasoning identifies evidence relevant to the answer.}
\label{fig:qualitative-more}
\end{figure*}

Figure~\ref{fig:qualitative-more} shows three examples that \method answers correctly: an in-domain X-ray, an OOD X-ray and a held-out endoscopy image. They serve as illustrations, not as an estimate of how often such cases occur.

\section{Baseline Protocols}
\label{app:baselines}

We compare with Qwen3-VL-8B, InternVL3-8B, MedGemma-4B and -27B, HuatuoGPT-Vision-7B (Qwen2.5-VL version) and -34B, Lingshu-7B, HealthGPT-Pro-8B, Fleming-VL-8B, MediX-R1-8B and MedVLThinker-7B. All models receive the same images and preprocessing, decode greedily and are scored by the same scorer. Forcing answer tags on direct-answer models costs them heavily in format, so besides the shared reasoning prompt we also report results with each model's own prompt. Under this protocol, the first line of a direct-answer model's response is taken as its answer, and reasoning models keep their released templates. MedVLThinker scores higher with our prompt than with its own template, so we use our prompt for it under both protocols.

\paragraph{Excluded RL systems.} We do not include MedCCO~\citep{medcco} or Med-R1~\citep{medr1}. MedCCO had released no checkpoint when we ran our evaluation, and its reported results use refined datasets and a different protocol. Med-R1 trains a separate checkpoint per modality or task on an 80/20 split of OmniMedVQA and reports multiple-choice letter accuracy, so OmniMedVQA is not held out for it.

\paragraph{Excluded MedMO.} We also leave out MedMO-8B~\citep{medmo}. It omits the required answer tags on over 95\% of the VQA-Med 2020 and Kvasir-VQA questions, so it cannot be scored consistently under the shared protocol. Its training corpus also includes VQA-RAD, SLAKE, PathVQA and PMC-VQA, which it uses as evaluation benchmarks, and the paper does not say whether only their training splits were used.

\paragraph{ViSurf.} As in the original method~\citep{visurf}, the gold answer is added to each group of eight rollouts as a ninth member and scored with the same rewards. Advantages are computed over all nine members, and the gold reward is replaced by the rollout mean whenever some rollout scores at least as high. All other settings, including the KL anchor and reward normalization, are those of \method, so the two runs differ only in how they use the gold answer. Without reasoning annotations the injected answer has no reasoning, and ViSurf's \emph{eliminate} rule cannot stop the policy from learning to skip it. The gold answer keeps a positive advantage mostly in all-wrong groups, its few tokens dominate the token-averaged loss, and the share of rollouts with valid reasoning drops from 74.5\% to under 5\% by step 70.

\paragraph{HPT switch.} The original HPT~\citep{lv2025hpt} switches each question between SFT on expert traces and RL. Since our data have no traces, our version supervises the gold answer after the rollout's own reasoning, as \method does.

\paragraph{RFT.} We follow the standard single-round recipe. The base model samples 16 answers per question, and the correct samples with valid reasoning (at most four per question) are used to train a new rank-32 LoRA for two epochs. Only 26.4\% of the open-ended questions are answered correctly at least once, compared with 86.5\% of the yes/no questions.

\section{LLM Judge and Human Audit}
\label{app:judge}
\label{app:human}

The LLM judge is DeepSeek-V4.1-Flash (temperature 0, thinking disabled; API model \texttt{deepseek-\allowbreak flash}), which decides whether a prediction and the gold answer mean the same thing. We judge each distinct pair once, do not send predictions longer than six words, and fall back to exact match for yes/no questions and for pairs where only one side is negated. Before the physician audit, we added the judge errors found during development (on projection, hierarchy and specificity) to a fixed blocklist.

\begin{table}[t]
\centering
\footnotesize
\setlength{\tabcolsep}{2.0pt}
\begin{tabular}{@{}l r rr@{}}
\toprule
\textbf{Sampling stratum} & $n$ & \textbf{Consensus pairs} & \textbf{Judge agrees} \\
\midrule
Judge same & 70 & 67 (95.7\%) & 66/67 (98.5\%) \\
Near different & 60 & 46 (76.7\%) & 36/46 (78.3\%) \\
Random different & 70 & 66 (94.3\%) & 66/66 (100.0\%) \\
\midrule
All & 200 & 179 (89.5\%) & 168/179 (93.9\%) \\
\bottomrule
\end{tabular}
\caption{\textbf{Blinded two-physician audit.} A pair counts as a consensus pair if both physicians gave it the same label other than \emph{unsure}. The strata were sampled on purpose, so the pooled agreement rate is not an estimate for the full population of pairs.}
\label{tab:human}
\end{table}

\paragraph{Two-physician audit.} Two physicians independently labeled 200 reference-candidate pairs in an interface that showed the image, the question and both answers, but not the judge's verdict or rationale, the model or the sampling stratum. We sampled 70 pairs that the judge called \emph{same}, 60 deliberately difficult near-differences and 70 random differences. The physicians gave the same label to 179 of the 200 pairs (89.5\%; Cohen's $\kappa{=}0.79$ with \emph{unsure} as a third label). One physician marked six pairs as \emph{unsure}, and on the other 194 pairs the two agreed on 179 (92.3\%; $\kappa{=}0.84$).

On the 179 pairs where the physicians agreed, the judge matched their label on 168 (93.9\%; 95\% Wilson CI 89.3--96.5\%; $\kappa{=}0.87$). For judged accuracy, false positives matter most: of the 67 \emph{same} verdicts with a physician consensus, both physicians accepted 66, so only one was a false positive. The other ten disagreements were all \emph{different} verdicts on near-difference pairs that the physicians judged equivalent. As Table~\ref{tab:human} shows, the judge is reliable on clear equivalences and clear differences, and its errors concentrate on borderline cases of specificity and granularity. Since the OOD open-ended ranking is also the same under EM and the judge, the gains of \method do not come from lenient semantic scoring. We still report both metrics because agreement drops to 78.3\% on the deliberately ambiguous near-differences.

\end{document}